\documentclass[preprint]{elsarticle}

\usepackage{multirow}
\usepackage{graphicx}
\usepackage{amsmath}
\usepackage{hyperref}
\usepackage{relsize}

\usepackage{bm}
\usepackage{amssymb}
\usepackage[utf8]{inputenc}
\usepackage{complexity}
\usepackage[linesnumbered,boxed]{algorithm2e}
\usepackage{endnotes}
\usepackage{listings}
\usepackage{mathtools}
\usepackage{verbatim}
\usepackage{mathrsfs}
\usepackage{longtable}
\usepackage{enumitem}
\usepackage{etoolbox}
\usepackage{float}
\usepackage{color}
\usepackage{tablefootnote}
\usepackage{threeparttable}
\usepackage{subcaption}
\usepackage[color]{changebar}

\cbcolor{red}

\journal{Swarm and Evolutionary Computation}

\newtheorem{example}{Example}

\usepackage{calc} 
\usepackage{tikz}
\usetikzlibrary{calc}

\makeatletter
\newcommand{\tikzmarkMath}[2]{%
	\tikz[%
	remember picture, 
	baseline = (#1.base),
	inner sep = 0pt,
	outer sep = 0pt, 
	] \node (#1) {$\m@th\displaystyle #2$};%
}
\makeatother

\begin{document}

\begin{frontmatter}
\title{Techniques for Inferring Context-Free Lindenmayer Systems With Genetic Algorithm}


\author[1]{Jason Bernard}
\ead{jason.bernard@usask.ca}
\author[1]{Ian McQuillan}
\ead{mcquillan@cs.usask.ca}

\address[1]{Department of Computer Science, University of Saskatchewan, Saskatoon, Canada}

\begin{abstract}
		Lindenmayer systems (L-systems) are a formal grammar system, where the most notable feature is a set of rewriting rules that are used to replace every symbol in a string in parallel; by repeating this process, a sequence of strings is produced. Some symbols in the strings may be interpreted as instructions for simulation software. Thus, the sequence can be used to model the steps of a process. Currently, creating an L-system for a specific process is done by hand by experts through much effort. The inductive inference problem attempts to infer an L-system from such a sequence of strings generated by an unknown system; this can be thought of as an intermediate step to inferring from a sequence of images. This paper evaluates and analyzes different genetic algorithm encoding schemes and mathematical properties for the L-system inductive inference problem. A new tool, the Plant Model Inference Tool for Deterministic Context-Free L-systems (PMIT-D0L) is implemented based on these techniques. PMIT-D0L is successfully evaluated on $28$ known L-systems created by experts with alphabets up to $31$ symbols, and PMIT-D0L can successfully infer even the largest of these L-systems in less than a few seconds. It is also evaluated and can correctly infer any system in a larger test set of algorithmically created L-systems with much larger alphabets.
\end{abstract}


\end{frontmatter}
	
\section{Introduction}
	Lindenmayer systems (L-systems), introduced in \cite{lindenmayer_lsystems}, are a bio-inspired grammar system that produces self-similar patterns that appear frequently in nature and especially in plants \cite{beauty}. L-systems produce a sequence of strings, where a string is obtained by the parallel application of rewriting rules to the previous string. Certain symbols can be interpreted as instructions to create images, and therefore a sequence of strings can describe a temporal process, which can be visually simulated by software such as the ``virtual laboratory'' (vlab) \cite{algorithmicbotany}. Such simulations can incorporate different geometries \cite{beauty,nishida1980k0l,godin_quantifying}, environmental factors \cite{drp_peach,watanabe_rice}, and mechanistic controls \cite{drp_auxin2009,nishida1980k0l}, and as such are useful for simulating plants. L-systems often consist of small textual descriptions that require little storage compared to real imagery. Certainly also, they have a low cost in currency, time, and labor to simulate \textit{in silico} compared to actually growing a plant, and realistic imagery can be produced with a well-constructed L-system.

\begin{figure}
	\centering
	\begin{subfigure}{.45\textwidth}
		\centering
		\includegraphics[height=3.5cm,keepaspectratio]{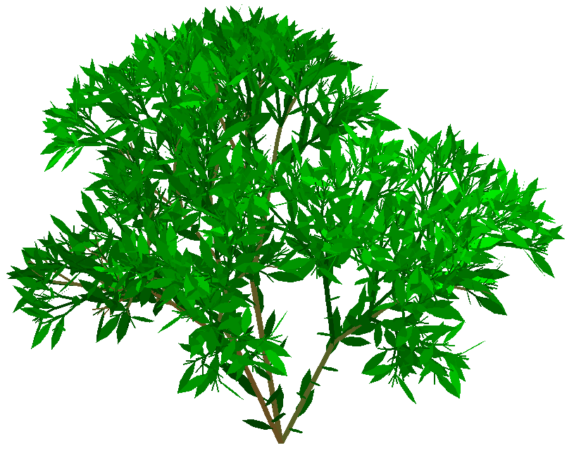}
		\caption{Fibonacci Bush after 7 generations as produced using vlab \cite{beauty}}
		\label{fig:1}
	\end{subfigure}
	\begin{subfigure}{.45\textwidth}
		\centering
		\includegraphics[height=3.5cm,keepaspectratio]{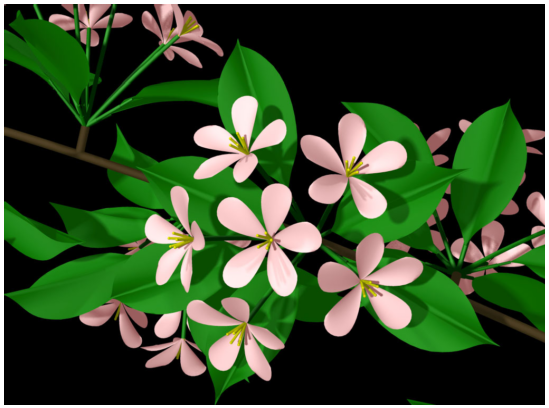}
		\caption{Apple twig with blossoms as produced using vlab \cite{beauty}}
		\label{fig:2}
	\end{subfigure}
\end{figure}

Formally, L-systems are described by an ordered tuple $G=(V,\omega,P)$ consisting of an alphabet $V$ (a finite set of allowed symbols), an axiom $\omega$ that is a string over $V$, and a finite set of productions, or rewriting rules, $P$. A deterministic context-free L-system (D0L-system) has rules of the form $A \rightarrow x$, where $A \in V$ is called the predecessor, $x$ is a string over V that is called the successor of $A$, with exactly one rule for each $A \in V$ as predecessor. Table \ref{table:results4} gives an example of a specific D0L-system created in \cite{morelli1991system}. Given a string $\omega_{i} = A_{1}\cdots A_{m}$ where each $A_{i} \in V, 1 \le i \le m$, a derivation step $\Rightarrow$ is defined by $A_{1} \cdots A_{m} \Rightarrow x_{1} \cdots x_{m}$ where $A_{i} \rightarrow x_{i}$ is in $P$, for each $i$, $1 \leq i \leq m$. When this process is repeated $n-1$ times starting with the axiom ($\omega = \omega_{1} \Rightarrow \omega_{2} \Rightarrow \cdots \Rightarrow \omega_{n}$), the sequence ($\omega_{1},\ldots,\omega_{n}$) is called the length-$n$ developmental sequence. By treating certain symbols as instructions for positioning and drawing in a 3D space (described in Section \ref{section:background}) temporal processes can be simulated using simulation software. Figures \ref{fig:1} and \ref{fig:2} show some structures built with D0L-systems in this fashion.

A difficult challenge is to determine an L-system that can accurately simulate a specific process, for example, modeling plant growth. In practice, this often involves manual measurements over time, scientific knowledge, and is done by hand by experts \cite{beauty,drp_modeling_by_hand,godin_quantifying,nishida1980k0l}. Although this approach has been successful, it does have notable drawbacks. Producing a system manually requires an expert, who are in limited supply, and it does not scale to producing arbitrarily many (perhaps closely related) models. Indeed, the manual process currently has been described as requiring ``tedious and intricate handwork'' \cite{galarreta_s0lbloodvessel} that could be improved if an automatic approach could ``infer rules and parameters automatically from real \ldots images'' \cite{galarreta_s0lbloodvessel}. Furthermore, when constructed manually, the more complex plant models require \textit{a priori} knowledge of the underlying mechanics of the plant. In contrast, inferring L-systems automatically may be used to reveal scientific principles of the underlying process, or as stated by Godin and Ferraro, automatic inference ``could be further exploited in combination with investigations at a biomolecular level to better understand plant development.'' \cite{godin_quantifying}

The ultimate goal of this line of research is to automatically determine an L-system from a sequence of plant images over time. There have been simplified variants of this problem that have been attempted thus far in the literature. One approach is to develop a tool as an aide for the expert to reduce the work load \cite{jacob_inferflowers,mock_wildwood}. With such approaches, the expert operator guides the tool towards a desirable model. Another approach is to build an automated method to convert a sequence of images directly to an L-system \cite{benes}. Yet another approach is to divide the problem into two separate steps, with the first step being an accurate segmentation of the images into a sequence of descriptive strings such that an L-system simulator  would draw an approximation of the input images; and the second step is to infer the L-system from the sequence of strings. This second step essentially involves taking a sequence of strings as input, and determining the unknown L-system which could give this sequence as its developmental sequence. This is known as the inductive inference problem and it dates back to early work on L-systems studied from the perspective of decidability \cite{rozenberg_systems}. We will focus exclusively on this approach. For the first step, this type of plant image segmentation used on a temporal image sequence has been studied separately, e.g.\ \cite{maize,tala_segmentation}. However, for inductive inference to be truly crucial in combination with image segmentation, it would need to be both fast and accurate so that it could be expanded to work with realistic complications such as noisy images, imperfect data, and other mechanisms built into L-systems. Existing work on inference and inductive inference of L-systems is described in Section \ref{section:existing}. Most previous attempts to infer L-systems implemented in the literature have involved only a single string as input rather than a sequence of strings \cite{nakano_inferD0Lerrorfree, runqiang_inferGA}; with a single string, the goal is to find the correct L-system that generated this string at some point during its computation. Certainly, there are fewer L-systems that could possibly generate a given sequence of strings versus a given single string within the sequence. Therefore, it could be substantially easier both computationally and in terms of accuracy to infer the correct L-system from a sequence of strings, thereby motivating the study of this problem.

This paper creates the Plant Model Inference Tool for Deterministic Context-Free L-systems (PMIT-D0L) \cite{bernard_pmitdv3,bernard_pmitml} that aims to be an automated approach to solve the inductive inference problem for D0L-systems. Towards that goal, PMIT-D0L uses a genetic algorithm (GA) to search for an L-system that produces a sequence of strings provided as input. In general, GAs search solution spaces in accordance with the encoding scheme used for the problem, and to-date most existing approaches to L-system inference use similar encoding schemes. This paper presents and analyzes different encoding schemes, both existing and novel, to show which are most effective for inferring L-systems. Additionally, some mathematical properties are used to shrink the solution space.

Between the encoding schemes and the use of mathematical properties based on necessary conditions, PMIT-D0L is able to infer all L-systems in a test suite of 28 previously developed L-systems where the number of letters is up to $31$ symbols; whereas, other approaches implemented in the literature are limited to $2$ symbols as described in Section \ref{section:existing}. The best encoding scheme, based on the novel approach of searching through allowable combinations of production {\it lengths} rather than productions directly, took no longer than $3.192$ seconds for each L-system, and it took $0.391$ seconds on average. All L-systems were inferred with $100\%$ accuracy with all encoding schemes. This is notable as the GA is being used to essentially learn the simulations from data. In addition, to test how PMIT-D0L works on larger L-systems, we algorithmically create a large set of D0L-systems while varying alphabet size. And indeed, PMIT-D0L was able to infer all D0L-systems tested even with up to 134 letters in at most one minute, which is far larger than any L-system in the literature.

There are many future directions required in order to fully realize automatic inference of L-systems. Although many modern L-systems produced by experts use additional features, especially rules that have parameters (parameterized L-systems \cite{beauty}), creating an inductive inference procedure for D0L-systems that is both fast and accurate is a big step forward. Firstly, it shows that the problem of inference from sequences of strings (and ultimately, images) has promise in the quest to determine a correct L-system versus other approaches. The additional data of a sequence of strings provides significantly more data than a single string to unambiguously recover a correct L-system in a fast way, and sequence data can be often practically obtained. Secondly, many uses of parameters in parameterized L-system rules behave like context-free L-systems during certain sections of their derivations (e.g. if the parameters are being used to incorporate a timing mechanism \cite{beauty,prusinkiewicz1990visualization}). The techniques developed in PMIT-D0L can be used for these sections, and can also be used to detect deviations corresponding to a change in the program via parameter. Therefore, studying D0L-system inference scientifically, and inferring D0L-systems in a fashion which can be extended into rules with parameters, is an important step towards the main long term objective. Indeed, PMIT-D0L provides both a fast and accurate implementation of inductive inference that is necessary for L-system inference from images, and is the first inductive inference implementation to do so.

The remainder of this paper is structured as follows. Section \ref{section:background} will describe some existing automated approaches for inferring L-systems. Section \ref{section:methodology} will discuss the different encoding schemes that can be used with PMIT-D0L, along with techniques for reducing the search space size, etc. Section \ref{section:results} discusses the data set, performance metrics, and the results of the evaluation of PMIT-D0L. Finally, Section \ref{section:conclusions} concludes the work and discusses future directions.

\section{Background}\label{section:background}

This section describes useful contextual and background information relevant to understanding this paper. It starts with describing some notation used. Since a GA is used as the search mechanism for this work, it contains a brief description of them. The section concludes with a discussion of some existing approaches to L-system inference.

\subsection{Notation}
\label{sec:notation}

An alphabet is a finite set of symbols. Given an alphabet $V$, a string (or word) over $V$ is any finite sequence of letters $A_{1}A_{2}\cdots A_{n}$, $A_{i}  \in V, 1 \leq i \leq n$. The set of all words over $V$ is denoted by $V^{*}$, which contains the empty string denoted by $\lambda$. Given a word $x \in V^{*}$, $|x|$ is the length of $x$, and $|x|_{B}$ is the number of $B$'s in $x$, where $B \in V$. Given $V = \{B_{1},\ldots,B_{k}\}$, the Parikh vector of a string $x \in V^{*}$ is $(|x|_{B_{1}},\ldots,|x|_{B_{k}})$. 

Given two words $x,y \in V^{*}$, $x$ is a subword of $y$ if $y = uxv$, for some $u,v \in V^{*}$ and in this case $y$ is said to be a superstring of $x$. Also, $x$ is a prefix of $y$ if $y=xv$ for some $v$, and $x$ is a suffix of $y$ if $y=ux$ for some $u$.

Given a D0L-system $G=(V,\omega,P)$ as defined in Section $1$, the successor of $A$ is indicated by $succ(A)$. Given a rewriting rule $A \rightarrow succ(A)$, and $B \in V$, then the number of symbols $B$ in $succ(A)$ is called the growth of $B$ by $A$, denoted by $M(A,B)$. These values are stored in a $|V| \times |V|$ matrix called the growth matrix $M(G)$ \cite{beauty}. Commonly, $V$ includes symbols to provide simple graphical instructions to simulation software (such as vlab \cite{algorithmicbotany}). One commonly used such instruction set is the ``Turtle Graphics'' \cite{beauty}. It is imagined as manipulating a turtle through a 2D or 3D space with a pen on its back. The turtle has a state consisting of its position and orientation. The symbols ``F'' and ``f'' move the turtle forward along its current orientation with the pen on or off respectively. In 2D, the symbols ``\textbf{+}'' and ``$\bm{-}$'' turn the turtle a predefined number of degrees left or right. In 3D, additional symbols are needed for pitch (``\&'' down and ``\textasciicircum'' up), and roll (``\textbackslash'' left and ``/'' right) \cite{beauty}.
 For branching processes, the symbols ``['' and ``]'' are used to start and stop a branch, which is implemented as pushing and popping the turtle's state on a stack and switching to it. It is usually the case that the symbols ``['', ``]'', ``\textbf{+}'', ``$\bm{-}$'' have identity productions. There are some instances where ``F'' may not have an identity production (e.g. some of the variants of ``Fractal Plant'' \cite{beauty}). Given a sequence of $n$ words $\rho$ over $V$, $G$ is said to be compatible with $\rho$ if $\rho$ is $G$'s length-$n$ developmental sequence. To differentiate the turtle graphic symbols ``\textbf{+}'', ``$\bm{-}$'' from the corresponding mathematical operators $+$ and $-$, the turtle graphics symbols will appear in bold as \textbf{+} and $\bm{-}$.

\subsection{Background on Genetic Algorithm}\label{section:GA}

The GA is described as follows by B\"ack \cite{back_geneticalgorithm}. The GA is an optimization algorithm based on evolutionary principles used to efficiently search $N$-dimensional (usually) bounded spaces. An encoding scheme is applied to convert a problem's solution space into one describable by a virtual genome consisting of $N$ genes. Each gene can be either a binary, integer, or real value and represents, in a problem specific way, an element of the solution to the problem. While there exists several types of value encoding schemes, a literal encoding directly represents an element of the solution to a problem. An example of a literal encoding scheme uses gene values to represent the length of a successor, so a value of $3$ indicates a successor length of $3$. In contrast, a mapped encoding could instead use a real value from $0$ to $1$ subdivided into sections that represent the different possible solutions.

In evolutionary biology, increasingly fit offspring can be created over successive generations by intermixing the genes of parents. Similarly, a GA functions by iterating over the selection, crossover, mutation, and survival operators until at least one termination condition (e.g.\ a time limit) is met \cite{back_geneticalgorithm}. There exist different types of these operators; however, this paper will describe only those used here. The function of the GA is controlled by the parameters: population size ($P$), crossover weight ($C$), and mutation weight ($M$). Prior to the first iteration, a GA first produces an initial population of $P$ random solutions. The selection operator chooses some number of pairs of genomes from the population using a selection technique. One such technique, a roulette wheel, is one where the chance of any option being selected (in this case a genome) is proportional to an associated value (in this case, the genome's fitness). For each pair of genomes, the crossover operator swaps a random selection of genes between them, resulting in $P$ child genomes. The chance for any gene to be swapped is equal to $C$. The mutation operator changes a random selection of genes to a random valid value in each child genome. The chance of any individual gene being mutated is equal to $M$. The child genomes are added to the population, and the population is culled to size $P$, thereby keeping the most fit genomes (elite survival). 

With PMIT-D0L, the following changes are made to the standard GA to encourage additional exploration. Although an individual genome may be selected for more than one pair, the same pair may not be selected more than once. If any genome has been modified by neither the crossover operator nor the mutation operator, then one gene is selected and mutated to ensure that at least one change has taken place. Where a mapped encoding is used, it is possible for two different genomes to map to the same solution. To prevent such solutions from dominating the population, genomes that map to the same solution are automatically culled (similarly during initialization, duplicated solutions are not permitted in the population).

\subsection{Existing Automated Approaches to L-system Inference}\label{section:existing}

Various approaches to L-system inference were surveyed in \cite{ben_naoum_surveryLsystems}. Here, only certain works most closely related to PMIT-D0L are described. There are several different broad approaches towards the problem: building by hand \cite{beauty,drp_modeling_by_hand,nishida1980k0l,godin_quantifying}, algebraic approaches \cite{doucet_algebra, nakano_inferD0Lerrorfree}, using mathematical properties \cite{nakano_inferD0Lerrorfree}, and search approaches \cite{runqiang_inferGA}.

Inductive inference was studied theoretically (without implementation) by Hermann and Rozenberg \cite{rozenberg_systems}, and Doucet \cite{doucet_algebra}. 
In \cite{rozenberg_systems}, the problem was studied from the perspective of decidability.
In \cite{doucet_algebra}, Doucet defined a matrix equation to simplify the problem. Let $\rho = (\omega_1, \ldots, \omega_n)$ be
input words over an $n-1$ letter alphabet, with the goal of finding a D0L-system that has $\rho$ as its length-$n$ developmental sequence. Let $Y$ be the $(n-1)\times (n-1)$ matrix such that row $i$ is the Parikh
vector of $\omega_i$ for $i$, $1 \leq i \leq n-1$, and let $Z$ be the $(n-1)\times (n-1)$ matrix such that row $i$ is the Parikh
vector of $\omega_{i+1}$ for $i$, $1 \leq i \leq n-1$. From the definition of matrix multiplication (also discussed in both \cite{doucet_algebra,mcquillan_poly}), if a D0L-system $G$ has $\rho$ as its length-$n$ developmental sequence,
then $YM(G) = Z$; that is, the unknown growth matrix must be a solution to the equation $YX = Z$. That said, there can be solutions to this equation that are not growth matrices of D0L-systems generating $\rho$. However, it is possible to search only solutions to this equation rather than arbitrary successor lengths in order to find a D0L-system that generates $\rho$. Indeed, for a given solution to this equation, it is straightforward and efficient to find an L-system with it as growth matrix if it exists, as described below in Section \ref{section:scanning}. Recently, this theoretical algorithm was extended to work for context-sensitive L-systems \cite{mcquillan_poly}. 
We implement a similar approach here as one encoding scheme under the name {\em row reduced matrix encoding}.
A somewhat similar approach \cite{doucet_algebra} was implemented with a tool called LGIN \cite{nakano_inferD0Lerrorfree} that infers L-systems from only a single string. LGIN looks exhaustively at the successor combinations, extracted from a single string in a developmental sequence that fulfills these equations. Since only a single string is used, the problem they are solving is more difficult than the inductive inference problem we are addressing, and indeed it does not guarantee to find a unique solution. LGIN only provides specific algorithms for one and two symbol alphabets (not including turtle graphic symbols), with larger alphabets described as ``immensely complicated'' \cite{nakano_inferD0Lerrorfree} and they are not described algorithmically; however, LGIN was evaluated on six variants of ``Fractal Plant'' \cite{beauty} and it was very fast having a peak time to find the L-system of less than one second for 5 of the 6 variants, and four seconds for the remaining variant.

\begin{figure}
	\centering
	\includegraphics[width=0.3\linewidth]{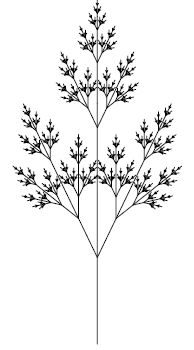}
	\caption{``Fractal Plant'' variant $\#5$ \cite{beauty}}
	\label{fig:3}
\end{figure}

Runqiang et al. \cite{runqiang_inferGA} investigated inferring an L-system directly from a description of a single image (essentially one string) using a GA. In their approach, they encode each symbol within the successors as a gene. The fitness function attempts to match the candidate system to the input string. Their approach is limited to an alphabet size of $2$ symbols and has a maximum combined length of all successors of $14$. Their approach is 100\% successful for variants of ``Fractal Plant'' \cite{beauty} with $|V| = 1$, and a 66\% success rate for variants of ``Fractal Plant'' as in Figure \ref{fig:3} with $|V| = 2$. Although they do not list any timings, their GA converged after a maximum of 97 generations, which suggests a short runtime. The main difference between their work and ours is they do not take a sequence of strings as input. They use an encoding scheme that we call \textit{ordered sequence of symbols} (discussed further in Section \ref{section:defining}), which we also implement and investigate here (on sequences of strings as input) which we compare to other encoding schemes.

\subsection{Scanning for Successors}\label{section:scanning}

A technique for finding productions based on the successor lengths for every $A \in V$ was previously described in \cite{mcquillan_poly}, which we call the \textit{scanning process}. This technique is used extensively in this research, and is described as follows. With L-systems, although the symbols are replaced in parallel, the sequence of successors in the new word is unchanged from the sequence of the original symbols; i.e., if $\omega_{i} = A_{1}A_{2} \ldots A_{m}$ with $A_{i} \in V$, $1 \le i \le m$, then $\omega_{i+1} = succ(A_{1})succ(A_{2}) \cdots succ(A_{m})$. Consider the case of $A_{1}$ in $\omega_{i}$. To find $succ(A_{1})$ it is only necessary to know the length of $succ(A_{1})$. If $|succ(A_{1})|$ is known (or different values for it are tested by searching), then the successor is the first $|succ(A_{1})|$ characters of $\omega_{i+1}$. The process of taking the next $l$ symbols (where $l$ is a hypothetical successor length) may be repeated for every new instance of a symbol encountered while scanning each word of a developmental sequence until every successor has been found, as described in \cite{mcquillan_poly}. With this fast algorithm, the goal is therefore to find a list of successor lengths that results in an L-system compatible with a developmental sequence; however, this may be done in a few ways. Most directly, a list of successor length combinations may be found by searching. Somewhat indirectly, the growth values for every $A,B \in V$ may instead be found, and a successor length for each $A \in V$ computed by summing the growth values for every $B \in V$. 

\section{Methodology}\label{section:methodology}

This section describes the design and procedure of PMIT-D0L. First, a high level overview of PMIT-D0L will be described, as this helps to contextualize the remainder of the section. Then, the techniques used to reduce the size of the defined search space are discussed.  This is followed by a description of the different encoding schemes used to define and search the space to infer an L-system (the different encoding schemes are evaluated separately to see which work best). The final two sub-sections describe the process used to optimize the control parameters of the GA, and finally, the fitness function and termination conditions.

PMIT-D0L makes two assumptions regarding L-systems to be inferred: no successor is the empty word, and that for branching L-systems, the branching symbols [ and ] are properly nested within each production (this is a common assumption, e.g.\ \cite{runqiang_inferGA} as improperly nested branching symbols are biologically meaningless). Most L-systems in the literature satisfy these assumptions. 

As mentioned in Section \ref{sec:notation}, some symbols, such as the turtle interpretation symbols, are usually created to have identity productions. Hence, a set $C \subset V$ of {\em constants} is provided as input where it is assumed that all symbols in $C$ have identity productions. This separation is commonly done with inference procedures (e.g.\ \cite{nakano_inferD0Lerrorfree} which separates out constants from the rest of the alphabet). In our experiments, we use the turtle interpretation symbols as the elements of $C$. Their known successors can speed up searches. In addition, let $\overline V = V - C$ (those symbols with unknown productions).

The pseudocode description of PMIT-D0L is provided in Algorithm \ref{algo:1}. This algorithm has been implemented in C++ with CLR extensions. It takes as input a sequence of strings $\rho = (\omega_{1}, \ldots ,\omega_{n}), \omega_{i} \in V^{*}, 1 \le i \le n$, an encoding scheme $En$ (the various encoding schemes are described in Section \ref{section:defining}), and a list of constants $C$. For turtle interpretation, we use the fixed ordering of the elements of $C$: $[,],+,-$, followed
by the symbols for yaw, pitch, roll, turn $180^{\circ}$ (if they are present), then $F,f$.
PMIT-D0L will return either a D0L-system compatible with $\rho$ or return that no D0L-system was found (which either means that one does not exist, or the GA terminated without finding a solution, as a GA cannot guarantee to find a solution) that is compatible with $\rho$. 

At a high level, the algorithm first attempts to find partial solutions by removing all constants from $\rho$, and then it searches for solutions to this sequence of strings using the GA. For all solutions found by the GA with a fitness of $0$, it adds them to a queue $Q$ (initialized in line \ref{initQ}) containing all partial solutions. Then, for each partial solution in the queue, it adds back in the first constant and tries determine, for each occurrence of this constant in $\rho$, which position of this constant in the next string must be produced from it (using a procedure described in detail in Section \ref{section:symbolprojection}). Then, from each of these partial solutions, it uses the GA to search for new partial solutions extended from it. It proceeds similarly for the second constant, etc., until
it has added back in all constants. It keeps track of the current constant to be added in to each partial solution with an index $currentConstant$ into
the list of constants $C$ (this value gets enqueued together with the partial solutions). One can see this basic procedure
with the while loop on line \ref{outerloop}, which removes all constants starting at index $currentConstant$ from $\rho$ (line \ref{removeconstants}); uses the GA to search the search space (line \ref{searchspace}); adds all new partial solutions found with a fitness of $0$ to $Q$;
takes one partial solution at a time from the queue (line \ref{dequeue}); and if it has already included all constants, then it is a complete solution and the L-system has been found and is returned; otherwise, it adds in the next constant (line \ref{dequeue}). The detailed procedure for adding one constant at a time to setup associations between consecutive strings is described in Section \ref{section:symbolprojection}.

The GA is searched in accordance with the desired encoding scheme $En$ (described in Section \ref{section:defining}). 
Before searching the search space with the GA, two approaches are used to reduce the size of the search space (executed in line \ref{refinelengthinfirst} before the loop, and within the loop in line \ref{refinelengthinloop} each time before searching with the GA). As each of the two techniques can lead to reductions using the other technique, the two methods are executed alternatingly in a loop as part of this pseudocode line until there are no changes.

The first reduction technique is done by examining possible successor length combinations and growth matrix values. Reducing the range of possible production lengths is crucial, as is evident from the scanning process described in Section \ref{section:scanning}, which can infer an L-system quickly and correctly from only the successor lengths (and successor lengths can be calculated from growth matrix values). Hence, narrowing down the range of possible successor lengths or growth matrix values can significantly reduce the size of the search space. Thus, the first step is to initialize several programming variables in line \ref{initState} that keep track of lower and upper bounds on each of these, which get refined as additional data is scanned and deductions are made. For each $A \in V$, the current lower bound (respectively upper bound) on the successor length of $A$ is denoted by $A_{\min}$ (respectively $A_{\max}$), and $A_{\min}$ is initialized to be $1$ as there are no erasing productions. The current upper and lower bounds for the growth of $B$ by $A$ for every pair of $A,B \in V$ is denoted by $(A,B)_{\min}$ and $(A,B)_{\max}$, respectively.  Since all letters in $C$ have a known identity production, we initialize 
for each symbol $T \in C$, $T_{\min} = T_{\max} = 1$,$(T,T)_{\min} = (T,T)_{\max} = 1$ and $(T,A)_{\min} = (T,A)_{\max} = 0$ for every $A \in V, A \neq T$. The technique to reduce the search space by refining these values is described in detail below in Section \ref{section:refininggrowth}.

The second reduction technique can be obtained using a concept called {\em successor fragments}. Given $A \in \overline{V}$, a word $\omega$ is an $A$-subword (respectively, $A$-prefix, $A$-suffix, $A$-superstring) if $\omega$
is a subword (respectively, prefix, suffix, superstring) of $succ(A)$. It is helpful to maintain, for each $A \in \overline{V}$,
words that we have deduced must be an $A$-subword, and similarly for
$A$-prefix, $A$-suffix, and $A$-superstring. It is evident that if there are multiple
$A$-prefixes, then only the longest is of interest as the shorter ones
are prefixes of the longest one, and similarly with the suffix. Hence, for each $A \in \overline{V}$, Algorithm \ref{algo:1} keeps strings $Sub_A$, $Pre_A$, and $Suf_A$ which stores the longest known $A$-subword,
$A$-prefix, and $A$-suffix respectively, and $Sup_A$ is the shortest known $A$-superstring. Also in line \ref{initState}, all strings are initialized to be empty.
The detailed method to help determine these fragments is given in Section \ref{section:refiningfragments}.

\begin{algorithm}[h]
	\SetAlgoLined
	\KwData{A sequence of strings $\rho$ over $V$, an encoding scheme $En$, and a list of symbols with known identity productions $C$}
	\KwResult{D0L-system compatible with $\rho$, or that no compatible D0L-system is found}
	$Q \longleftarrow \emptyset$ // initialize queue of solutions\; \label{initQ}
	// initialize state consisting of length and growth bound variables, fragment successors, and $currentConstant \longleftarrow 0$\; \label{initState}
	refine length and growth bounds, and fragment successors\; \label{refinelengthinfirst}	
	\While{$currentConstant \leq |C|$}{\label{outerloop}
		$\rho'$ $\longleftarrow$ remove constants starting at $currentConstant$ from $\rho$\;  \label{removeconstants}
		refine length and growth bounds, and fragment successors\; \label{refinelengthinloop}	
		search $space$ based on $En$ (Sec.\ \ref{section:defining}) and add each solution found with fitness $0$ together with state to $Q$\; \label{searchspace}
		\eIf{$Q$ is not empty}{
			dequeue $solution$ and state and make state current\;\label{dequeue}	
			\eIf{$currentConstant = |C|$}{
				return $solution$\;
			}{
				add letter $C[currentConstant]$ to $solution$ (Sec.\ \ref{section:symbolprojection})\; \label{addnext}
				increase $currentConstant$\;
			}
		}{return ``none found''\;}
	}
	\caption{High-level D0L-system inference procedure.}\label{algo:1}
\end{algorithm}

\subsection{Search Space Reduction}\label{section:reducing}

In this section, the techniques that are used by PMIT-D0L to reduce the size of the solution space (with any of the encoding schemes) using mathematical properties of D0L-systems will be described. As previously mentioned, these are applied in Algorithm \ref{algo:1} in line \ref{refinelengthinfirst} and \ref{refinelengthinloop} before each GA search, and the two techniques are alternatively executed in a loop until there are no more changes. Being based on necessary conditions guarantees that all valid solutions are in the remaining search space (if there is a D0L-system that can generate the input strings).

\subsubsection{Refining Successor Relationships}\label{section:refiningfragments}

As PMIT-D0L runs, it can determine additional successor fragments, which can help in turn to reduce growth bounds. Certain prefix and suffix fragments can be found for the first and last symbols in each input word by the following process.  Consider two words such that $\omega_{i} \Rightarrow \omega_{i+1}$. It is possible to scan $\omega_{i}$ from left-to-right until the first symbol from $\overline V$ is scanned (say, $A$, where the word scanned is $\alpha A$). Then, in $\omega_{i+1}$, PMIT-D0L skips over the symbols of $C$ in $\alpha$ (since each symbol in $\alpha$ has a known identity production), and the next $A_{\min}$ symbols (the current lower bound for $|succ(A)|$), $\beta$ say, must be an A-prefix fragment. Furthermore, since branching symbols must be well-nested within a successor, if a $[$ symbol is met, the prefix fragment must also contain all symbols until the matching $]$ symbol is met. Similarly, an A-superstring fragment can be found by skipping $\alpha$, then taking the next $A_{\max}$ symbols from $\omega_{i+1}$ (the upper bound on $|succ(A)|$). If a superstring fragment contains a $[$ symbol without the matching $]$ symbol, then it is reduced to the symbol before the unmatched $[$ symbol. The lower and upper bounds $(A,B)_{\min}$ and $(A,B)_{\max}$ for each $B \in V$ can be then possibly improved by counting the number of $B$ symbols in any prefix and superstring fragments respectively. For a suffix fragment, the process is identical except the scan goes from right-to-left starting at the end of $\omega_{i}$.

\begin{example}
	Consider input strings:
	\begin{eqnarray*}
	&\omega_{1} = \textbf{+}\textbf{+}\textbf{+}A[\bm{-}FF][\textbf{+}F]BF&\\
	&\omega_{2} = \textbf{+}\textbf{+}\textbf{+}\textbf{+}A[\bm{-}FF][\bm{-}FF][\textbf{+}F][\textbf{+}F]BFF\text{.}&
	\end{eqnarray*} 
	Assume thus far PMIT-D0L has calculated that $A_{\min} = 2$ and $A_{\max} = 8$. It can scan $\omega_{1}$ until $A$ is found and record that $\alpha = \textbf{+++}$. An $A$-prefix fragment is $\textbf{+}A$ as those are the first two ($A_{\min}$) symbols of $\omega_{2}$ after skipping $\alpha$. An $A$-superstring fragment is $\textbf{+}A[-FF][$ as those are the first eight ($A_{\max}$) symbols of $\omega_{2}$ after skipping $\alpha$, which can be reduced to $\textbf{+}A[\bm{-}FF]$ due to the unbalanced $[$ symbol. By counting within the prefix fragment, lower bounds on the growth for $A$ are $(A,\textbf{+})_{\min} := 1$ and $(A,A)_{\min} := 1$, while upper bounds can be found from the superstring fragment to be $(A,\textbf{+})_{\max} := 1, (A,\bm{-})_{\max} := 1$, $(A,A)_{\max} := 1$,$(A,[)_{\max} := 1$, $(A,])_{\max} := 1$ and $(A,F)_{\max} := 2$.
	\label{example:1}
\end{example}	


If a symbol has a known successor, then it may be possible under certain circumstances to ``line it up'' with the symbol(s) it produces. In such circumstances, the derivation is sliced into two parts, and this reduces the possible productions for the symbols in each part. To illustrate the concept, consider the simple example:
\begin{equation}
\nonumber
\begin{split}
\omega_{1}\text{:}\ A&\tikzmarkMath{A1}{\textbf{+}}B\\[1.5em]
\omega_{2}\text{:} \underbrace{ABA}_{\text{\normalsize succ(A)}}&\tikzmarkMath{B1}{\textbf{+}}\underbrace{BBB}_{\text{\normalsize succ(B)}} \\[0em]
\begin{tikzpicture}[overlay,remember picture,distance=0.0cm]
\draw[thick,->,black,shorten >=2pt,shorten <=2pt] (A1.south) to (B1.north);
\end{tikzpicture}
\end{split}
\end{equation}
As $\textbf{+} \in C$, it must be that $\textbf{+} \rightarrow \textbf{+}$ is a production, and the $\textbf{+}$ symbol in $\omega_{1}$ may only produce the $\textbf{+}$ in $\omega_{2}$. This splits the derivation such that everything to the left of the $\textbf{+}$ in $\omega_{1}$ must produce everything to the left of the $\textbf{+}$ in $\omega_{2}$, i.e.\ $A \rightarrow ABA$ must be a production. Similarly to the right of the $\textbf{+}$, $B \rightarrow BBB$ must be a production.

In practice, it is unusual for a single position of one string to be uniquely associated with a position in the next string, as in the example. More often, any individual position may be associated to multiple positions of the next word. However, a sequence of symbols, each with known successor, may be unique. For example, in the string $A[\textbf{+}B][\bm{-}B]A[\textbf{+}[\bm{-}C]][\bm{-}[\textbf{+}D]]$, the
individual symbols $[$, $]$, $\textbf{+}$, and $\bm{-}$ alone might not uniquely associate to their successors; however, a sequence of symbols such as $][\bm{-}$ or $]][\bm{-}[\textbf{+}$ are potentially unique, and as such may be associated to a unique location. To make use of this observation, for each word, a list of possible associations between every position of a symbol in $C$ to positions of the next word is constructed. This list of associations is called a \textit{marker map}. A marker map is constructed based on both individual symbols and sequences of symbols, which are referred to as \textit{candidate markers}. Associating a candidate marker to potential successors takes into account that a number of symbols must be reserved for symbols that appear before and after the candidate marker. For example, if $\omega_{1} = A\textbf{+}BC\bm{-}, \omega_{2} = A\textbf{+}BC\textbf{+}C\bm{-}$, then $\textbf{+}$ associates with both $\textbf{+}$'s in $\omega_{2}$, both are candidate markers. But since $B_{\min} + C_{\min} + \bm{-}_{\min} \geq 3$ (as there are no empty word successors) the final $3$ symbols of $\omega_{1}$ produce at least the $\textbf{+}C\bm{-}$ of $\omega_{2}$. This eliminates the
second $\textbf{+}$ in $\omega_{2}$ as being produced by the $\textbf{+}$ in
$\omega_{1}$, and the $\textbf{+}$ in $\omega_{1}$ can only be associated to the first $\textbf{+}$. If, following the construction of the marker map, a candidate marker is not uniquely associated with its successor, then it is removed from the marker map.

Once the marker map has been calculated, it can help significantly to improve the length bounds and successor fragments. Consider a derivation $\omega_{i} \Rightarrow \omega_{i+1}$ expanded as
\setlength{\arraycolsep}{0.0em}
\begin{equation*}\omega_{i,1}A_{1}\omega_{i,2}\cdots A_{m} \omega_{i,m+1} \Rightarrow \omega_{i+1,1}succ(A_{1})\omega_{i+1,2}\cdots succ(A_{m}) \omega_{i+1,m+1},
\label{equation:omega}
\end{equation*}
where each $A_{j}$, $1 \leq j \leq m$ in $\omega_{i}$ has already been associated to the annotated successor in $\omega_{i+1}$ forming a marker. It follows that $\omega_{i,j} \Rightarrow \omega_{i+1,j}$ for all $j$, $1 \le j \le m+1$. Indeed, from this, improved successor fragments, growth and length bounds may be found.

\subsubsection{Refining Growth and Length Bounds}\label{section:refininggrowth}

Here the bounds on $(A,B)_{\min}$ and $(A,B)_{\max}$ are improved. As all properties are run in a loop, these bounds are also influenced by successor fragments, and markers as described above, which can result in significantly improved bounds versus just examining what can be determined from Parikh vectors alone. A programming variable for the \textit{accounted for growth} of a symbol $A \in V$ for $2 \leq i \leq n$, denoted as $G_{acc}(i,A)$ is:
$$G_{acc}(i,A) := \sum_{B \in V} (|\omega_{i-1}|_{B} \cdot (B,A)_{\min})\text{.}$$
The \textit{unaccounted for growth} for a symbol $A$, denoted as $G_{ua}(i,A)$, is computed as $G_{ua}(i,A) := |\omega_{i}|_{A} - G_{acc}(i,A)$. 

The unaccounted for growth can be used to improve the growth bounds. In particular, $(B,A)_{\max}$ is set (if it can be reduced) under the assumption that all unaccounted for $A$ symbols are produced by $B$ symbols. Furthermore, $(B,A)_{\max}$ is set to be the lowest such value computed for any word from $2$ to $n$, where $B$ occurs, as any of the $n-1$ words can be used to improve the maximum. And, $|succ(B)|_{A}$ must be less than or equal to $(B,A)_{\min}$ plus the additional unaccounted for growth of $A$ divided by the number of $B$ symbols (if there is at least one) in the previous word, as computed by
$$(B,A)_{\max} := \min_{2 \leq i \leq n, \atop |\omega_{i-1}|_{B} > 0} \Bigg((B,A)_{\min} + \left \lfloor \tfrac{G_{ua}(i,A)} {|\omega_{i-1}|_{B}} \right \rfloor \Bigg)\text{.}$$
Indeed, the accounted for growth of $A$ is always updated whenever values of $(B,A)_{\min}$ change, and the floor function is used since $|succ(B)|_{A}$ is a non-negative integer. For example, if $\omega_{i-1} = ABA$, $\omega_{i} = ABABBBABA$, $(A,A)_{\min} = 1$, and $(B,A)_{\min} = 0$, then the accounted for growth of $A$ in $\omega_{i}$ is computed by $G_{acc}(i,A) = (A,A)_{\min} \cdot |\omega_{i-1}|_{A} + (B,A)_{\min} \cdot |\omega_{i-1}|_{B} = 1 \cdot 2 + 0 \cdot 1 = 2$. This leaves two $A$'s in $\omega_{i}$ unaccounted for. An upper bound on the value of $|succ(A)|_{A}$ is set when the $A$'s in $\omega_{i-1}$ produce all of the unaccounted for growth in $\omega_{i}$. So $A$ produces its minimum ($(A,A)_{\min} = 1$) plus the unaccounted for growth of $A$ in $\omega_{i}$ ($2$) divided by the number of $A$'s in $\omega_{i-1}$ ($|\omega_{i-1}|_{A} = 2$), hence $(A,A)_{\max} := 2$. Similarly, $(B,A)_{\max}$ is achieved when only $B$'s produce all unaccounted for growth of $A$; this sets $(B,A)_{\max}$ to $(B,A)_{\min} = 0$ plus the unaccounted for growth ($2$) divided by the number of $B$'s in $\omega_{i-1}$ ($1$), which is 2.

Once $(B,A)_{\max}$ has been determined for every $A,B  \in V$, the observed words are re-processed to compute possibly improved values for $(B,A)_{\min}$. Indeed for each $(B,A)$, if $x := \sum_{C \in V \atop C \neq B} (C,A)_{\max}$, and $x < |\omega_{i}|_{A}$, then this means that $|succ(B)|_{A}$ must be at least $\left \lceil \tfrac{|\omega_{i}|_{A} - x} {|\omega_{i-1}|_{B}} \right \rceil$, and then $(B,A)_{\min}$ can be set to this value if its bound is improved. For example, if $\omega_{i-1}$ has 2 $A$'s and 1 $B$, and $\omega_{i}$ has 10 $A$'s, and $(A,A)_{\max} = 4$, then at most two $A$'s produce eight $A$'s, thus one $B$ produces at least two $A$'s ($10$ total minus $8$ produced at most by $A$), and $(B,A)_{\min}$ can be set to $2$.

In a similar fashion, the length bounds $A_{\min}$ and $A_{\max}$ can be set using unaccounted for length.

\subsubsection{Solution Projections}\label{section:symbolprojection}

As previously defined, $C \subset V$ where all symbols in $C$ have a known identity production, and $\overline V = V - C$. Since a symbol in $\overline V$ cannot be produced by a symbol in $C$, in the while loop of Algorithm \ref{algo:1}, it is possible to first infer an L-system over reduced alphabets with fewer constants, and add one constant at a time. For example, if $V = \{A,B,C,\text{[},\text{]},\textbf{+},\bm{-}\}$ and $C$ is a list of constants $(\text{[},\text{]},\textbf{+},\bm{-})$, then one can first find each successor of $A,B,C$ projected to $\{A,B,C\}$. After solving this initial problem, then a series of problems are solved adding in each symbol of $C$ to determine where it belongs in each successor. Note that ``['' and ``]'' are completed together due to the assumption that they are properly nested. Overall, symbol filtering simplifies the inference problem by allowing for an iteration of lining up constants between consecutive words. Although more searches are needed, they are each in a smaller search space. 

As described above, PMIT-D0L removes the symbols of $C$ temporarily by projecting $\rho$ onto a reduced alphabet, and then it iteratively re-adds each symbol of $C$ back into the problem one at a time. Let a solution to one of these reduced problems be called a \textit{partial solution}, as it partly describes the final successors. The process for using the partial solutions towards the next partial solution by adding in the next symbol of $C$ is conceptually similar to that used for markers, as positions in pairs of consecutive words will be ``lined up'', and from there, successor relationships deduced. To describe this, some terminology for this process is provided. For every derivation step $\omega_{i} \Rightarrow \omega_{i+1}$, every position in $\omega_{i}$ is scanned, and associated to the possible locations in $\omega_{i+1}$ that it must produce. Let
$\omega_{i+1}'$ be equal to $\omega_{i+1}$ but with all symbols of $C$ erased.
As the algorithm proceeds, a position $j$ of $\omega_i$ is said to be {\em certain} if 
positions $k,l$ have been determined such that letter $D$ at position $j$ of $\omega_i$ must produce exactly the symbols of $\omega_{i+1}$ between positions $j$ and $k$; and it is {\em uncertain} otherwise. If a position is certain, this means it has already been determined exactly the positions that this specific $D$ must produce in the next word (which also determines the successor of this letter). Note that if position $j$ is labelled by an element of $C$, then $k =l$ and position $k$ of $\omega_{i+1}$ must be $D$ since $D$ is a constant. 
This property is purely programmatic, and a position can change from uncertain to certain as the algorithm proceeds. Similarly, a position $j$ of $\omega_i$ labelled by some symbol $D \in \overline{V}$ is said to be {\em projected-certain} if
positions $k,l$ have been determined such that letter $D$ at position $j$ of $\omega_i$ must produce exactly the symbols of $\omega_{i+1}'$ between positions $j$ and $k$; and it is {\em projected-uncertain} otherwise.
This means, a position $j$ is projected-certain if it has been determined exactly which positions of $\omega_{i+1}$ can be produced
by this position, ignoring adjacent positions labelled by symbols of $C$.

As we iterate the loop of Algorithm \ref{algo:1}, every time we re-add a new symbol $D \in C$  in line \ref{addnext}, this involves scanning every position of each string of $\rho$ to try to determine whether they are certain. It is often possible to determine that certain positions are certain if adjacent positions are certain. We will describe the idea first with an example.
Consider $\omega_1$ and $\omega_2$ in Equation \ref{equation:4} to \ref{equation:6}, and assume that it has already been determined that the successor of $A$ projected to $\overline{V}$ is $BAB$, and the successor of $B$ projected to $\overline{V}$ is $AB$.
That means that, by scanning $\omega_1$ from left-to-right, each position in $\omega_1$ labelled by a symbol of $\overline{V}$
is projected-certain, as the first $A$ must produce the first three symbols of $\omega_2'$, the $B$ must produce the next two symbols, and the second $A$ must produce the last three symbols. This also indicates that the first $A$ must produce $B[\textbf{+}A]B$ (labelled $\alpha$ in Equation \ref{equation:4}), but it is not yet clear what adjacent symbols of $C$ can also be produced by this $A$ (so this $A$ is not yet certain). Similarly, the $B$ must also produce $A{\bm{-}}[\textbf{+}B]$ of $\omega_2$, but it is unclear as to which adjacent symbols of $C$ are produced by $B$. Lastly, the final $A$ must produce the final $B[\textbf{+}A]B$ of $\omega_2$ but adjacent symbols of $C$ are unclear. 

Next, the loop of Algorithm \ref{algo:1} re-adds the $[$ and $]$ symbols as the first two constants. As these symbols do not occur in $\omega_1$, there are no new symbols to line up and the procedure does not do anything. The next letter of $C$ it re-adds is the $\textbf{+}$ symbol. The $\textbf{+}$ symbol is uncertain at this stage,
as the $\textbf{+}$ could produce either of the two annotated $\textbf{+}$'s in $\omega_2$ as shown in Equation \ref{equation:4}; it cannot produce the first $\textbf{+}$ as the first $A$ must map to positions that include this symbol.
Even though the \textbf{+} is uncertain at this stage, the word $B[\textbf{+}A]B$ ($\alpha$ in Equation \ref{equation:4}) can be declared an $A$-prefix. The next letter re-added is $\textbf{-}$, and it is concluded that the $\textbf{-}$ is certain, as it has already been determined that the final $A$ maps to the final $B[\textbf{+}A]B$ with perhaps some adjacent symbols of $C$ to the left, and therefore, it cannot map to the second last $\textbf{-}$. Since $\textbf{-}$ is now certain, this implies that the final $A$ is certain which indicates that the successor of $A$ is $B[\textbf{+}A]B$. This implies that the first $A$ must map to exactly the first $B[\textbf{+}A]B$ and the first $A$ is now certain, which also resolves the certainty of the first $\textbf{+}$, and lastly it implies that the $B$ is certain as well and produces exactly
$\textbf{+}A{\bm{-}}[\textbf{+}B]$. Thus, the L-system is determined.
\begin{equation}
\begin{split}
&	\omega_{1}\text{:} A\tikzmarkMath{A1}{\textbf{+}}B\bm{-}A\\[1.5em]
&	\omega_{2}\text{:} \underbrace{B\text{[}\textbf{+}A\text{]}B}_{\alpha}\hspace{-0.50mm}\tikzmarkMath{B1}{\textbf{+}}\tikzmarkMath{B2}{\textbf{+}}A\bm{-}\text{[}\textbf{+}B\text{]}\bm{-}B\text{[}\textbf{+}A\text{]}B \\
\begin{tikzpicture}[overlay,remember picture,distance=0.3cm]
\draw[thick,->,black,shorten >=2pt,shorten <=2pt] (A1.south) to[out=-90,in=90,looseness=400] (B1.north);
\draw[thick,->,black,shorten >=2pt,shorten <=2pt] (A1.south) to[out=-90,in=90,looseness=400] (B2.north);	
\end{tikzpicture}
\end{split}
\label{equation:4}
\end{equation}
\begin{equation}
\begin{split}
&\omega_{1}\text{:} A\tikzmarkMath{A1}{\textbf{+}}B\tikzmarkMath{A2}{\bm{-}}A\\[1.5em]
&\omega_{2}\text{:} B\text{[}\textbf{+}A\text{]}B\tikzmarkMath{B1}{\textbf{+}}\tikzmarkMath{B2}{\textbf{+}}\hspace{-0.50mm}\underbrace{A\bm{-}\text{[}\textbf{+}B\text{]}}_{\beta}\hspace{-0.30mm}\tikzmarkMath{B3}{\bm{-}}B\text{[}\textbf{+}A\text{]}B \\
\begin{tikzpicture}[overlay,remember picture,distance=0.3cm]
\draw[thick,->,black,shorten >=2pt,shorten <=2pt] (A1.south) to[out=-90,in=90,looseness=400] (B1.north);
\draw[thick,->,black,shorten >=2pt,shorten <=2pt] (A1.south) to[out=-90,in=90,looseness=400] (B2.north);	
\draw[thick,->,black,shorten >=2pt,shorten <=2pt] (A2.south) to[out=-90,in=125,looseness=400] (B3.north);	
\end{tikzpicture}
\end{split}
\label{equation:5}
\end{equation}
\begin{equation}
\begin{split}
&\omega_{1}\text{:} A\tikzmarkMath{A1}{\textbf{+}}B\tikzmarkMath{A2}{\bm{-}}A\\[1.5em]
&\omega_{2}\text{:} B\text{[}\textbf{+}A\text{]}B\tikzmarkMath{B1}{\textbf{+}}\tikzmarkMath{B2}{\textbf{+}}A\bm{-}\text{[}\textbf{+}B\text{]}\tikzmarkMath{B3}{\bm{-}}\hspace{-0.75mm}\underbrace{B\text{[}\textbf{+}A\text{]}B}_{\text{\normalsize succ(A)}} \\
\begin{tikzpicture}[overlay,remember picture,distance=0.4cm]
\draw[thick,->,black,shorten >=1pt,shorten <=2pt] (A2.south) to[out=-90,in=90,looseness=400] (B3.north);	
\end{tikzpicture}
\end{split}
\label{equation:6}
\end{equation}
Hence, every time it executes line \ref{addnext} of Algorithm \ref{algo:1}, it scans each position from left-to-right of each string and assesses certainty of each position in this fashion; and it repeats this process in a loop until there are no more changes, so that it can fully consider how the certainty of positions can change as the certainty of other positions change.

\subsection{Defining and Searching the Search Space}\label{section:defining}

This section describes the different encoding schemes used in this research, and in some cases existing approaches to using encoding schemes \cite{runqiang_inferGA, doucet_algebra,nakano_inferD0Lerrorfree}, for inferring D0L-systems. Broadly, the encoding schemes can be broken down into three categories: ordered sequence of symbols (OSoS), growth-based, and length-based. The OSoS approaches take the viewpoint that a successor is an ordered sequence of unknown symbols, and so the search space is represented in this fashion. Another approach investigated in this research is to instead attempt to determine successor length combinations as the unknown, as an intermediate step, before determining the actual productions using the scanning process of Section \ref{section:scanning}. Similarly, the growth values may be inferred first, and then simply summed for each $A \in V$ to produce a successor length followed by the scanning process.

\subsubsection{Ordered Sequence of Symbols Encoding}

While the technique of building a search space based on the idea of searching for the symbol in each position of each successor has been previously investigated \cite{runqiang_inferGA,mock_wildwood}, PMIT-D0L creates this search space with additional requirements.

For every $A \in V$, a successor may be encoded as follows. For the remainder of this section, we create a special symbol $\overline \emptyset$. A number of genes equal to $A_{\max}$ is defined as this is the greatest number of symbols that may exist in the successor. The current values for $Pre_{A}$ and $Suf_{A}$ immediately identify a number of genes at the beginning and end equal to the length of the prefix and suffix respectively. For example, if $A_{\max} = 7$, $Pre_{A} = A$, and $Suf_{A} = BB$, then the genome would appear as follows A \rule{0.75em}{.5pt} \rule{0.75em}{.5pt} \rule{0.75em}{.5pt} \rule{0.75em}{.5pt} B B, where \rule{0.75em}{.5pt} represents an unknown symbol, which could be set to $\overline \emptyset$ if $|succ(A)| < 7$ (implying no symbol of $V$ exists in that position of this successor considered). Each of the genes are permitted to have a real-value from 0 to 1. Growth bounds can reduce the options however. To continue the previous example, if $(A,B)_{\max} = 2$, then since there are already two occurrences of $B$ symbols in the successor, $B$ needs not be a choice for any of the remaining genes. Minimum growth values can be similarly enforced. Further, if $(A,A)_{\min} = 3$, $(A,C)_{\min} = 2$, and the successor is A A \rule{0.75em}{.5pt} \rule{0.75em}{.5pt} \rule{0.75em}{.5pt} B B, then the remaining genes must be either $A$ or $C$ regardless of how many symbols are in $V$. The lower bound on successor length is enforced by making $\overline \emptyset$ unable to be selected until $A_{\min}$ symbols exist in the successor.

Furthermore, after a list of possible symbols for a gene is determined, instead of giving each symbol an equal chance of selection, the ranges can be improved based on frequency of letters (and letters with context) occurring in $\rho$. For example, if the choices for a gene are $A$ and $B$, then instead of setting the probabilities of $A$ and $B$ to $0.5$ for each gene, it is weighted by the frequency with which $A$ and $B$ occur in $\rho$. It is also possible to incorporate a context (sliding a window) to improve the probabilities. For example, with the string $AABAACAAB$, if the successor state is $A$ $A$ \rule{0.75em}{.5pt}, then  using two symbols of context within the string shows $AAB$ occurs twice and $AAC$ once, and therefore $B$ is given a $2/3$ probability and $A$ is given a $1/3$ probability. Note, all strings of $\rho$ are used to compute the associated probability. This encoding scheme is called OSoS($N$), where $N$ is the length of the context. This paper evaluates both OSoS(1) and OSoS(2).

\subsubsection{Growth Encoding}

This approach, called PMIT-D0L(G), searches using the GA within the computed lower and upper bounds for $M(A,B)$, of which there are $|V|^{2}$ values. PMIT-D0L(G) uses a literal encoding scheme and is similar to those seen in \cite{doucet_algebra, nakano_inferD0Lerrorfree}, where the correct value of a dimension is each value in $M(A,B)$. For each combination of growth matrix tested, the sum of each row is obtained to give a length and then the scanning process is used. 

The implementation chosen for this encoding allows for the possibility that a candidate does not satisfy the property that, for each $B \in V$, and for the growth values currently being assessed in $M'$, $\sum_{A \in V} (|\omega_{i}|_{A} M'(A,B)) = |\omega_{i+1}|_{B}, 1 \le i < n$ by choosing each gene independently from the values chosen for other genes. This avoids backtracking as the GA is free to select any values within the lower and upper bounds for each $M'(A,B)$. An alternate encoding scheme was also tested by adapting mapped ranges so that this length constraint equation needed to be satisfied. The results of an evaluation were found to be very similar to those for the encoding scheme described above, and so results for this alternative approach are omitted.

\subsubsection{Length Encoding}

This approach, called PMIT-D0L(L), uses the scanning process that requires a successor length for each $A \in V$. Each dimension is  mapped onto an integer value representing the length of a successor of a symbol in $V$. The values of each dimension represents the length of a successor, with the dimensions bounded by the computed upper and lower bounds for length. Compared to the growth-based approach, although the bounds on the individual dimensions are larger; i.e., the lower bound $A_{\min} \ge \sum_{B \in V} (A,B)_{\min}$, and the upper bound $A_{\max} \le \sum_{B \in V} (A,B)_{\max}$, with the length-based approach. The number of dimensions in the search space is $|V|$ with the length-based approach.


As with the growth-based approach, an alternative encoding was also implemented that enforced the constraint, for lengths $x_A$ for $A \in V$ currently being assessed, $\sum_{A \in V} (|\omega_{i}|_{A} x_A) = |\omega_{i+1}|, 1 \le i < n$. The evaluation showed that the results were also not significantly different overall, and so this approach is not discussed further.

\subsubsection{Row-Reduced Matrix Encoding}

Recall that Doucet \cite{doucet_algebra} recognized
that the productions could be represented as a matrix equation which is given in Section \ref{section:existing}. A similar approach was implemented. Let $Y$ be the matrix where each row $i$, from $1$ to $n-1$ is the Parikh vector of $\omega_{i}$, and let $Z$ be the matrix where each row is a Parikh vector of $\omega_{2}$ to $\omega_{n}$. In this case, if $M$ is a growth matrix of an D0L-system with $\rho$ as its length-$n$ developmental sequence, then $YM = Z$ is true. In Doucet's original work, they proposed to solve for $M$, and where $Y$ is invertible, $Y^{-1}Z$ is a unique solution, and if the solution is not unique, to use linear Diophantine equations.

It is also possible to replace $M$ with the length of each production, called the \textit{successor length matrix}, and $Z$ is replaced with the column vector consisting of the length of $\omega_{2}$ to $\omega_{n}$, and this modified equation $Y X = Z$ must have the length of the productions of any D0L-system having $\rho$ as developmental sequence as a solution. We can try all solutions to this equation (of which there can be more solutions than the correct L-system) similarly to the growth matrix version of this equation, and test whether each solution gives a D0L-system compatible with $\rho$.
The remainder of this discussion will be presented in the context of a length-based matrix; however, similar logic applies to a growth-based matrix by replacing growth values for successor lengths. Furthermore, Gaussian elimination can be applied to this equation in order to produce parameterized equations in terms of successor lengths. That is, after Gaussian elimination is applied, it results in a set of linear Diophantine equations, where the successor lengths are the variables, e.g.\ $5 X_{1} + 3 X_{2} = 24$. This would mean that $5$ times the first successor length plus $3$ times the second successor length must be $24$. It is easier to search the space defined by solutions to these equations than searching all possible length combinations. Each successor length only gets substituted for variables that appear in exactly one equation. In these cases, there are an infinite number of possible solutions over the integers, however when inferring L-systems, the successor lengths are constrained to be natural
numbers and within the bounds on the lengths provided by the lengths of the words in $\rho$, and it is therefore finite. For each equation, the encoding scheme used to search for a solution has $N$ genes, where $N$ is the number of variables in an equation. The range of values for each gene is $A_{\min}$ to $A_{\max}$ for the symbol $A$ the gene is representing (which can be more restricted than solutions to Diophantine equations due to the additional mathematical properties in Section \ref{section:reducing} used that takes the sequences of the words into account). This encoding scheme is designated as PMIT-D0L(M+L) to indicate the addition of the matrix operation. For the matrix based on growth values, it is designated as PMIT-D0L(M+G).

\subsection{Parameter Optimization}\label{section:parameteroptimization}

As described in Section \ref{section:GA}, the function of the GA is controlled by the population size ($P$), crossover weight ($C$), and mutation weight ($M$) parameters. Optimizing these parameters is difficult based on general principles, since the optimal settings will depend on the characteristics of the fitness landscape, which is problem specific \cite{hyperparameter}. As such, typically, the parameters are set by doing a \textit{hyperparameter search}. Bergstra and Bengio \cite{hyperparameter} found that using a Random Search provides an effective means to optimize the GA's parameter settings. Using Random Search works as follows. A range of good values is selected for each control parameter. In this case, based on the work by Grefenstette \cite{grefenstette_optimalGAparameter}, the ranges were set to $10 \le P \le 125$ in increments of $5$, $0.6 \le C \le 0.95$ in increments of $0.05$, and $0.01 \le M \le 0.20$ in increments of 0.01, with additional values of $0.001$ and $0.0001$ permitted. An initial mid-range value is selected for each parameter ($P=60$,$C=0.8$,$M=0.10$), and this is considered the current parameter value set. Iteratively, sixteen trials of PMIT-D0L are executed with a random variant of the current parameter value set. Each parameter is randomly modified up or down by no more than two increments, i.e.\ $P$ may be modified by $-10$, $-5$, $0$, $+5$, $+10$, while also remaining within the ranges above. The variant parameter set that provides the best fitness value is considered the new current parameter value set. In the case of a tie, which was quite common with PMIT-D0L, the fastest execution time is used. When none of the sixteen trials provide an improvement over the current parameter value set, the hyperparameter search terminates. The resulting parameter value sets for each variant of PMIT-D0L is shown in Table \ref{table:parameter}.

\begin{table}
	\centering
	\begin{tabular}{|c||c|c|c|c|c|c|}
		\hline
		\multirow{2}{*}{Parameter} & \multicolumn{6}{c|}{PMIT-D0L} \\ \cline{2-7}
		& OSoS(1)           & OSoS(2) & G       & M+G    & L        & M+L    \\ \hline
		P & 110  & 105  & 90   & 95   & 105  & 100 \\ \hline 
		C & 0.80 & 0.80 & 0.85 & 0.90 & 0.85 & 0.85 \\ \hline 
		M &	0.17 & 0.14 & 0.07 & 0.09 & 0.10 & 0.10 \\ \hline 	
	\end{tabular}
	\caption{Optimized parameters for each variant of PMIT-D0L}
	\label{table:parameter}
\end{table}

\subsection{Fitness Function and Termination Conditions}\label{section:fitnessfunction}

After a candidate L-system $G$ is produced from the solution $S$, the following process is used to evaluate fitness. To begin, any $G$ which produces more than double the expected number of symbols is assigned the maximum fitness value so it (practically) guaranteed to be culled in the survival step. Starting with $\omega_{1}$, a developmental sequence of length $n$ is produced using $G$ denoted as $\overline \rho$. For each $\overline \omega_{i} \in \overline \rho$, or until it terminates (see below), the symbol in each position of $\omega_{i}$ is compared to the corresponding position in $\overline \omega_{i}$. An error is counted if the symbol does not match (like Hamming distance), or if there is no corresponding symbol (i.e., one of the strings is longer or shorter than the other). For example, when comparing $\omega_{i} = XYXXXY$ to $\overline \omega_{i} = XYYX$, there are four errors. The third and fourth symbols differ, and additionally $\omega_{i}$ has six symbols, while $\overline \omega_{i}$ has only four. This process terminates when the number of errors for the $i^{th}$ derivation is greater than zero, as any errors will cascade forward. The fitness value is computed as the number of errors divided by the number of expected symbols (e.g., $4/6$), plus the number of unchecked derivations ($n-i$). This encourages the GA to find solutions that incrementally match $\rho$.

PMIT-D0L uses a three-part termination condition to determine when to stop running. Ideally, PMIT-D0L terminates when a solution is found with a fitness value of 0.0 as this corresponds to an L-system that produces $\rho$ as its length $n$ developmental sequence. PMIT-D0L will also terminate when the population is considered to have converged to prevent the GA from acting as a random search and skewing the results. First, the current generation $Gen_{best}$ is recorded whenever a new best solution is found. If an additional $Gen_{best}$ generations pass without finding a new best solution, the population is considered converged. To prevent random chance from causing early termination, PMIT-D0L must process at least $1,000$ generations. PMIT-D0L also terminates after a time limit is reached. For this paper, the time limit was set to four hours; however, this was mainly used to control the overall experimental time. In practice, a user may be willing to wait less or more time to find an L-system.

\section{Evaluation}\label{section:results}

\subsection{Data Set}\label{section:data}

To evaluate PMIT-D0L's ability to infer D0L-systems, ten fractals, plus the six plant-like fractal variants (one shown in Figure \ref{fig:3}) inferred by the existing program LGIN \cite{beauty,nakano_inferD0Lerrorfree}, and twelve other biological models were selected from the vlab online repository \cite{algorithmicbotany}. The biological models consist of ten algaes, apple twig with blossoms (shown in Figure \ref{fig:2}), and a ``Fibonacci Bush'' (shown in Figure \ref{fig:1}). The dataset compares favourably to similar studies where only some variants of one or two models are considered \cite{nakano_inferD0Lerrorfree,runqiang_inferGA}. The data set is also of greater complexity by considering models with alphabets from between $2$ to $31$ (excluding constants) symbols compared to two symbol alphabets \cite{nakano_inferD0Lerrorfree,runqiang_inferGA}. In all, 28 D0L-systems that have been created manually by experts were included. An example of a larger L-system is given in Table \ref{table:results4}.
\begin{table}
	\centering
	\begin{tabular}{|l|}
		\hline
		\multicolumn{1}{|c|}{Productions} \\ \hline
		$c \rightarrow FFFz[\textbf{+}k][\bm{-}r]FFfd$\\ \hline
		$z \rightarrow Fz$\\ \hline
		$k \rightarrow lmfF$\\ \hline
		$r \rightarrow stfF$\\ \hline
		$d \rightarrow FFFz[\textbf{+}k][\bm{-}r]FFfe$\\ \hline
		$l \rightarrow fF$\\ \hline
		$m \rightarrow n$\\ \hline
		$s \rightarrow fF$\\ \hline
		$t \rightarrow u$\\ \hline
		$e \rightarrow FFFz[\textbf{+}fj]FFfg$\\ \hline
		$n \rightarrow fFF[\bm{-}A]Fo$\\ \hline
		$u \rightarrow fFF[\textbf{+}A]Fv$\\ \hline
		$j \rightarrow abF$\\ \hline
		$g \rightarrow FFFz[\textbf{+}k][\bm{-}r]FFfh$\\ \hline
		$A \rightarrow fFB$\\ \hline
		$o \rightarrow fFF[\bm{-}B]Fp$\\ \hline
		$v \rightarrow fFF[\textbf{+}B]Fw$\\ \hline
		$a \rightarrow Ff$\\ \hline
		$b \rightarrow c$\\ \hline
		$h \rightarrow FFFz[\textbf{+}k][\bm{-}r]FFfi$\\ \hline
		$B \rightarrow fFC$\\ \hline
		$p \rightarrow fFF[\bm{-}C]Fq$\\ \hline
		$w \rightarrow fFF[\textbf{+}C]Fx$\\ \hline
		$i \rightarrow FFFz[\bm{-}fj]FFfc$\\ \hline
		$C \rightarrow fFD$\\ \hline
		$q \rightarrow fFF[\bm{-}D]F$\\ \hline
		$x \rightarrow fFF[\textbf{+}D]F$\\ \hline
		$D \rightarrow fF$\\ \hline
	\end{tabular}
	\caption{L-system for \textit{Dipterosiphonia} v1 \cite{morelli1991system}.}
	\label{table:results4}
\end{table}

In order to further examine the performance of PMIT-D0L using a simple GA beyond the test set above, an additional set of D0L-systems is algorithmically created for various alphabet sizes (ignoring constants) $|\overline{V}|$ starting with $|\overline{V}| = 2$. For each size $|\overline{V}|$, a set of $100$ D0L-systems is procedurally generated. 

The design of the L-system generator was intended to be simple, while still generating realistic successors. Towards this goal, the successors of the expertly created L-systems were examined. Many of the fractal L-systems have successors where production and graphical symbols alternate. For example, Dragon Curve's productions are $X \rightarrow X\textbf{+}YF\textbf{+}$ and $Y\rightarrow \bm{-}F\bm{-}Y$. Also, the successors typically consist predominantly of a few symbols of $\overline{V}$ followed by a constant. Alternatively, they have long sequences of constants which make the problem easier for PMIT-D0L to infer due to the ability to line up constants. For example, \textit{Aphanocladia} has productions 
\begin{eqnarray*}
A & \rightarrow & BA, \hspace{1cm} B \rightarrow  U[\bm{-}C]UU[\textbf{+}/C/]U, \hspace{1cm} U \rightarrow ffFFFF,\\
C &\rightarrow & FFfFFfFFfFF[\bm{-}FFFF]fFFfFF[\textbf{+}FFF]fFFfFF[\bm{-}FF]fFFf.
\end{eqnarray*} 
Similar successors are found in \textit{Ditria reptans}, \textit{Ditria zonaricola}, \textit{Metamorphe} and others. Notice also that with the branching patterns, a directional symbol immediately follows a branch open symbol $[$, which makes sense otherwise it would continue inline with the preceding angle.

The following methodology is therefore used to procedurally generate D0L-systems.
The axiom is created by concatenating up to $4$ random symbols of $\overline{V}$. Each production is created by iteratively concatenating symbols until a randomly selected length is reached (with a caveat on length described below) where $10$ is used as an upper bound on production length. This bound of $10$ was chosen to be larger than the average production length (which is $7.8$) for expertly created D0L-systems and because we probabilistically discourage long stretches of constants. For each position of a production, there is a base $80\%$ chance of selecting a symbol of $\overline{V}$; this is reduced by $20\%$ for every consecutive symbol of $\overline{V}$ (e.g.\ if $ABA$ has been selected, then there is a $20\%$ chance that the next symbol will be a symbol from $\overline{V}$). Once it has chosen to use some constant (or some element of $\overline{V}$), the particular letter is chosen with equal probability. Branching patterns must comply with the following rules: 1) they must be properly nested within each production, and 2) enforcing that a direction change symbol occurs after a ``['' (essentially this is a type of normal form). The length of the successor may be exceeded to enforce the branching rules, i.e., additional ``]'' symbols can be added to make it properly nested. The L-system is validated by generating $|\overline{V}|+1$ strings for $\rho$ and confirming that every symbol in $\overline{V}$ occurs at least once within the first $\overline{V}$ strings of $\rho$; it is easy to show that this is equivalent to ensuring that every symbol of $\overline{V}$ could eventually be reached, and therefore the L-system does not contain useless symbols.

\subsection{Performance Metrics}

Two performance metrics are used to measure how well PMIT-D0L can infer D0L-systems. The first metric is \textit{success rate} (SR) which is defined as the percentage of times PMIT-D0L can find any L-system compatible with the input sequence. The second metric is \textit{mean time to solve} (MTTS), in seconds, but measured to the millisecond level since some models can be determined in sub-second time. Time was measured using a single core of an Intel 4770 @ 3.4 GHz with 12 GB of RAM on Windows 10. These metrics are consistent with those found in literature \cite{nakano_inferD0Lerrorfree, runqiang_inferGA}.

\subsection{Results}

The first set of results is MTTS results for the L-systems in the expert-created L-systems, shown in Table \ref{table:results1}. 
The size of the variables (excluding constants) is given in the second column, and the MTTS using the various encoding schemes is given in columns 3 through 8.
The SR was $100\%$ for all L-systems and encoding schemes and is not shown. For PMIT-D0L(M+G) and PMIT-D0L(M+L), the systems where the matrix was invertible are marked with ``*'' as no searching was required. An  average for each encoding technique is also provided. The lowest average was for PMIT-D0L(M+L).

\begin{figure}
	\centering
	\includegraphics[width=0.95\linewidth]{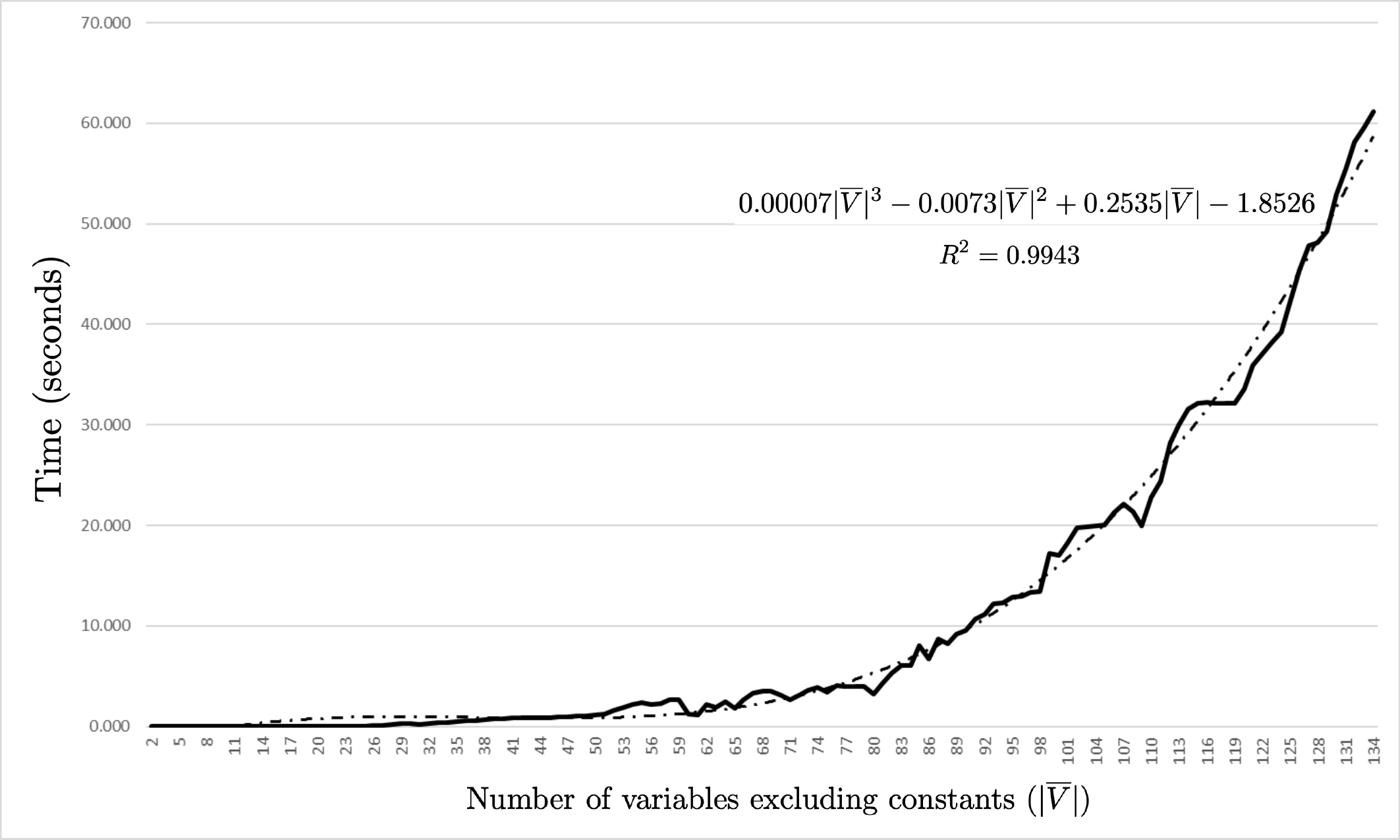}
	\caption{The solid line shows average MTTS across $100$ procedurally generated D0L-systems for each number of non-constant variables ($\overline{V}$) from 1 to 134. The dash line shows the polynomial trend line for the MTTS results, and the equation is provided.}
	\label{fig:graphvvstime}
\end{figure}

Figure \ref{fig:graphvvstime} shows $MTTS$ using PMIT-D0L(M+L) with the procedurally generated L-systems described in Section \ref{section:data}. The solid line gives the line graph of the average MTTS over 100 procedurally generated L-systems for each $|\overline{V}|$. This is tested for each size of $|\overline{V}|$ from $1$ to $134$, and all were inferred with $100\%$ accuracy in less than one minute. The polynomial trend line is also provided.

\begin{table*}[h]
	\centering
	\resizebox{\textwidth}{!}{%
	\begin{tabular}{|c||c|c|c|c|c|c|c|}
		\hline
		\multirow{2}{*}{Model} & \multirow{2}{*}{$|\overline{V}|$} & \multicolumn{6}{c|}{PMIT} \\ \cline{3-8}
		&   & OSoS(1)           & OSoS(2)        & G              & M+G             & L              & M+L    \\ \hline
		Algae                       & 2 & \textbf{0.001}    & \textbf{0.001} & \textbf{0.001} & \textbf{0.001*} & \textbf{0.001} & \textbf{0.001*} \\ \hline
		Cantor Dust                 & 2 & \textbf{0.001}    & \textbf{0.001} & \textbf{0.001} & \textbf{0.001*} & \textbf{0.001} & \textbf{0.001*} \\ \hline
		Dragon Curve                & 2 & \textbf{0.001}    & \textbf{0.001} & \textbf{0.001} & \textbf{0.001}  & \textbf{0.001} & \textbf{0.001}  \\ \hline
		E-Curve                     & 2 & 24.312            & 23.075         & 0.026          & \textbf{0.025}  & 0.088          & 0.029  \\ \hline
		Fractal Plant v1            & 2 & \textbf{0.001}    & \textbf{0.001} & \textbf{0.001} & \textbf{0.001*} & \textbf{0.001} & \textbf{0.001*} \\ \hline
		Fractal Plant v2            & 2 & \textbf{0.001}    & \textbf{0.001} & \textbf{0.001} & \textbf{0.001*} & \textbf{0.001} & \textbf{0.001*} \\ \hline
		Fractal Plant v3            & 2 & \textbf{0.001}    & \textbf{0.001} & \textbf{0.001} & \textbf{0.001*} & \textbf{0.001} & \textbf{0.001*}\\ \hline
		Fractal Plant v4            & 2 & \textbf{0.001}    & \textbf{0.001} & \textbf{0.001} & \textbf{0.001*} & 0.002          & \textbf{0.001*}\\ \hline
		Fractal Plant v5            & 2 & \textbf{0.001}    & \textbf{0.001} & \textbf{0.001} & \textbf{0.001*} & 0.003          & \textbf{0.001*}\\ \hline
		Fractal Plant v6            & 2 & \textbf{0.001}    & \textbf{0.001} & \textbf{0.001} & \textbf{0.001*} & 0.002          & \textbf{0.001*}\\ \hline
		Gosper Curve                & 2 & 0.022             & 0.023          & \textbf{0.001} & \textbf{0.001*} & 0.006          & \textbf{0.001*}\\ \hline
		Koch Curve                  & 1 & \textbf{0.001}    & \textbf{0.001} & \textbf{0.001} & \textbf{0.001*} & \textbf{0.001} & \textbf{0.001*}\\ \hline
		Peano                       & 2 & 0.236             & 0.235          & \textbf{0.202} & 0.210           & 5.916          & 0.221  \\ \hline
		Pythagoras Tree             & 2 & \textbf{0.001}    & \textbf{0.001} & \textbf{0.001} & \textbf{0.001*} & \textbf{0.001} & \textbf{0.001*}\\ \hline
		Sierpenski Triangle v1      & 2 & \textbf{0.001}    & \textbf{0.001} & \textbf{0.001} & \textbf{0.001*} & 0.010          & \textbf{0.001*}\\ \hline
		Sierpenski Triangle v2      & 2 & \textbf{0.001}    & \textbf{0.001} & \textbf{0.001} & \textbf{0.001*} & 0.003          & \textbf{0.001*}\\ \hline
		\textit{Aphanocladia}       & 4 & 0.004             & 0.004          & 0.005          & \textbf{0.001}  & 0.206          & 0.007  \\ \hline
		\textit{Dipterosiphonia} v1 & 28 & 11.885            & 10.871         & 123.008        & 3.820           & 178.718        & \textbf{1.639}  \\ \hline
		\textit{Dipterosiphonia} v2 & 5 & 0.278             & \textbf{0.236} & 0.348          & 1.114           & 1.055          & 1.199  \\ \hline
		\textit{Ditria reptans}     & 4 & 0.009             & 0.009          & 0.004          & \textbf{0.003}  & 0.039          & \textbf{0.003}  \\ \hline
		\textit{Ditria zonaricola}  & 4 & 0.012             & 0.010          & 0.006          & \textbf{0.003}  & 0.161          & 0.007  \\ \hline
		\textit{Herpopteros}        & 4 & 0.019             & 0.017          & 0.007          & \textbf{0.004}  & 0.070          & 0.006  \\ \hline
		\textit{Herposiphonia}      & 5 & 0.026             & 0.022          & 0.016          & \textbf{0.013}  & 0.190          & 0.015  \\ \hline
		\textit{Metamorphe}         & 5 & 7732.255          & 512.040        & 1.381          & 3.793           & \textbf{0.632} & 2.387  \\ \hline
		\textit{Pterocladiella}    & 30 & 22.631            & 8.805          & 0.944          & \textbf{0.881}  & 4.120          & 3.192  \\ \hline
		\textit{Tenuissimum}        & 31 & 0.851             & 0.871          & 0.603          & \textbf{0.520}  & 120.619        & 1.141  \\ \hline
		Apple Twig                  & 17 & 1.012             & 0.963          & \textbf{0.914} & \textbf{0.914}  & 0.957          & 0.970  \\ \hline
		Fibonacci Bush              & 8 & 500.262           & 112.332        & 4.095          & 37.525          & 8.663          & \textbf{0.108}  \\ \hline \hline
		Average                     & n/a & 296.208           & 23.912         & 4.699          & 1.744           & 11.481         & 0.391 \\ \hline
	\end{tabular}}
	\caption{Comparison of MTTS in seconds for different encoding schemes for PMIT-D0L with the best MTTS bolded. $|\overline{V}|$ indicates the number of constant symbols in the L-system. SR is $100\%$ for all executions. Results with ``*'' indicate an invertible matrix.}
	\label{table:results1}
\end{table*}

\subsection{Discussion}

It is evident from Table \ref{table:results1} and the average row that OSoS(1) and OSoS(2) are performing worse than the other encoding schemes, and therefore using some form of length with the scanning process seems to be best. However, OSoS(2) is faster overall than OSoS(1) (especially for \textit{Metamorphe}), and therefore additional context is helping with OSoS. Overall, PMIT-D0L is $100\%$ successful at inferring a diverse range of D0L-systems. The L-systems in the data set have different numbers of successors, with different lengths, and structures. With respect to using Doucet's \cite{doucet_algebra} approach to find a unique solution, the matrix is found to be invertible for a little less than half of the L-systems in the test set, but never for any of the biological models. However, for both PMIT-D0L(M+G) and PMIT-D0L(M+L) the addition of the matrix operation to reduce the search space provides a benefit over PMIT-D0L(G) and PMIT-D0L(L) respectively. It is not so clear cut as to which encoding scheme is best, although PMIT-D0L(M+L) is the fastest overall, finishing in an average of $0.391$ seconds. Certainly, it can be seen that PMIT-D0L(M+G) and PMIT-D0L(M+L) tend to be better than those without the matrix operations, but the timings tend to be close. However, for Fibonacci Bush, PMIT-D0L(M+L) performed much better than PMIT-D0L(M+G). Overall, both are quite fast, and the same can be said for all of the growth-based and length-based encoding schemes. This leads perhaps to the conclusion that choosing between a growth-based or length-based encoding scheme is not so important, but rather the success of PMIT-D0L (of any type) is largely attributed to the space reduction techniques.

Since one benefit of automatic L-system inference is to infer L-systems that are not easily found by experts, the performance of PMIT-D0L was evaluated against procedurally generated D0L-systems of increasing complexity. It was found that PMIT-D0L seems to exhibit polynomial behaviour with respect to the number of successors with a trend line of $0.00007 |\overline{V}|^3 - 0.0073 |\overline{V}|^2 + 0.2535 |\overline{V}| - 1.8526$.  Using just a simple GA, PMIT-D0L was able to infer D0L-systems with $|\overline{V}| \le 134$ in one minute or less. This is approximately five times larger than the largest D0L-system that could be found in the literature.

While all of the techniques are essential overall, the use of successor relationships extracts information by utilizing the sequence in which the symbols appear in $\rho$. This works by capitalizing on the fact that, even though the symbols are replaced in parallel, the order of the successors is the same as the symbols; i.e, if $\omega_{i} = A_{1}A_{2} \cdots A_{m}$ then $\omega_{i+1} = succ(A_{1})succ(A_{2}) \cdots succ(A_{m})$. Thus, if the relationship between $A_{j}$ and $succ(A_{j})$ is known (or even partially known), much can be deduced about the location of $succ(A_{j})$ in $\omega_{i+1}$ based on the location of $A_{j}$ in $\omega_{i}$. This in turn allows for the deduction of symbols near $A_{j}$. This is one of the main differences between PMIT-D0L and other existing approaches. Capturing information from the symbol sequence of strings in $\rho$ should provide guidance towards future investigations on inferring L-systems.

\section{Conclusions}\label{section:conclusions}

This paper presented an evaluation of different encoding schemes for the Plant Model Inference Tool for deterministic context-free L-systems (PMIT-D0L) to infer L-systems from a sequence of strings. Some of the encoding schemes are based on  modifications of earlier works, while some are novel. The classical encoding schemes look at the problem of inferring successors by letting each position of each production be an unknown \cite{runqiang_inferGA}. Here, we use a novel encoding scheme which considers the length of each production (or each Parikh vector) as an unknown, as it is straightforward to determine the L-system from the production lengths.

The evaluation of the different encoding schemes does not indicate a clear best encoding, however length-based approaches are the fastest for this test set. Much of this paper focused on techniques for reducing the search space size, using necessary conditions. 
Some of the techniques, such as setting up associations between constants (usually created from graphical symbols) are novel and particularly effective. The techniques are effective to the degree that the choice between growth-based and length-based is not particularly critical for this particular test set.

The inductive inference of L-systems from input strings allows for much more rapid development of models than the current approach of building models by hand \cite{beauty,drp_modeling_by_hand,godin_quantifying,nishida1980k0l}, which can take considerable effort. Additionally, by going directly from observation (strings) to a model, allows for mechanistic principles to be possibly revealed, as opposed to requiring expert knowledge to build the model.

Since PMIT-D0L seems capable of inferring L-systems with fairly large alphabets (at least $31$ symbols in the test set in a fast manner, and much larger in the algorithmically generated L-systems), this work will be used as a base for investigating the inference of other, more complex, L-system extensions such as stochastic L-systems, parametric L-systems, and for using images as input. While inferring parametric L-systems is more complex than D0L-systems, some of the principles and techniques from this paper may be applicable, especially if these L-systems behave like D0L-systems in certain parts of their derivation, such as when parameters are used as a timing mechanism. Also, the use of markers and solution projection should still be applicable as well to parametric L-systems. As a final note, the speed of PMIT-D0L can be further enhanced using parallel processing, which will be explored in the future.

\section{Acknowledgements}

This research was undertaken thanks in part to funding from the Canada First Research Excellence Fund, and the National Science Engineering Research Council (I. McQuillan grant 2016-06172, J. Bernard scholarship).

We thank the anonymous reviewers for valuable suggestions which considerably improved the presentation of the paper.

\bibliographystyle{elsarticle-num}
\bibliography{pmit_d0l}{}

\begin{thebibliography}{10}
\expandafter\ifx\csname url\endcsname\relax
  \def\url#1{\texttt{#1}}\fi
\expandafter\ifx\csname urlprefix\endcsname\relax\def\urlprefix{URL }\fi
\expandafter\ifx\csname href\endcsname\relax
  \def\href#1#2{#2} \def\path#1{#1}\fi

\bibitem{lindenmayer_lsystems}
A.~Lindenmayer, Mathematical models for cellular interaction in development,
  parts {I} and {II}, Journal of Theoretical Biology 18~(3) (1968) 280--315.

\bibitem{beauty}
P.~Prusinkiewicz, A.~Lindenmayer, The Algorithmic Beauty of Plants, Springer
  Science \& Business Media, 2012.

\bibitem{algorithmicbotany}
{University of Calgary}, \href{http://algorothmicbotany.org}{Algorithmic
  {B}otany}.
\newline\urlprefix\url{http://algorothmicbotany.org}

\bibitem{nishida1980k0l}
T.~Nishida, K0{L}-system simulating almost but not exactly the same
  development-case of {Japanese Cypress}, Memoirs of the Faculty of Science,
  Kyoto University, Series B 8~(1) (1980) 97--122.

\bibitem{godin_quantifying}
C.~Godin, P.~Ferraro, Quantifying the degree of self-nestedness of trees:
  application to the structural analysis of plants, IEEE/ACM Transactions on
  Computational Biology and Bioinformatics 7~(4) (2010) 688--703.

\bibitem{drp_peach}
M.~T. Allen, P.~Prusinkiewicz, T.~M. DeJong, Using {L}‐systems for modeling
  source–sink interactions, architecture and physiology of growing trees: The
  {L‐PEACH} model, New Phytologist 166~(3) (2005) 869--880.

\bibitem{watanabe_rice}
T.~Watanabe, J.~S. Hanan, P.~M. Room, T.~Hasegawa, H.~Nakagawa, W.~Takahashi,
  Rice morphogenesis and plant architecture: measurement, specification and the
  reconstruction of structural development by {3D} architectural modelling,
  Annals of {B}otany 95~(7) (2005) 1131--1143.

\bibitem{drp_auxin2009}
P.~Prusinkiewicz, S.~Crawford, R.~S. Smith, K.~Ljung, T.~Bennett, V.~Ongaro,
  O.~Leyser, Control of bud activation by an auxin transport switch,
  Proceedings of the National Academy of Sciences 106~(41) (2009) 17431--17436.

\bibitem{morelli1991system}
R.~Morelli, R.~Walde, E.~Akstin, C.~Schneider, L-system representation of
  speciation in the red algal genus {Dipterosiphonia} ({Ceramiales},
  {Rhodomelaceae}), Journal of Theoretical Biology 149~(4) (1991) 453--465.

\bibitem{drp_modeling_by_hand}
P.~Prusinkiewicz, L.~Mündermann, R.~Karwowski, B.~Lane, The use of positional
  information in the modeling of plants, in: Proceedings of the 28th Annual
  Conference on Computer Graphics and Interactive Techniques, ACM, 2001, pp.
  289--300.

\bibitem{galarreta_s0lbloodvessel}
M.~A. Galarreta-Valverde, M.~M. Macedo, C.~Mekkaoui, M.~Jackowski,
  Three-dimensional synthetic blood vessel generation using stochastic
  {L}-systems, in: Medical Imaging: Image Processing, 2013, p. 86691I.

\bibitem{jacob_inferflowers}
C.~Jacob, Genetic {L-system} programming: breeding and evolving artificial
  flowers with {Mathematica}, in: Proceedings of the First International
  Mathematica Symposium, 1995, pp. 215--222.

\bibitem{mock_wildwood}
K.~J. Mock, Wildwood: The evolution of {L}-system plants for virtual
  environments, in: Proceedings of the 1998 IEEE World Congress on
  Computational Intelligence, IEEE, 1998, pp. 476--480.

\bibitem{benes}
J.~Guo, H.~Jiang, B.~Benes, O.~Deussen, X.~Zhang, D.~Lischinski, H.~Huang,
  Inverse procedural modeling of branching structures by inferring {L}-systems,
  ACM Transactions on Graphics 39~(5) (2020).

\bibitem{rozenberg_systems}
G.~Herman, G.~Rozenberg, Developmental Systems and Languages, North-Holland,
  Amsterdam, 1975.

\bibitem{maize}
N.~Khan, O.~Lyon, M.~Eramian, I.~McQuillan, A novel technique combining image
  processing, plant development properties, and the {Hungarian} algorithm, to
  improve leaf detection in maize, in: 2020 IEEE/CVF Conference on Computer
  Vision and Pattern Recognition Workshops (CVPRW 2020), 2020, pp. 330--339.

\bibitem{tala_segmentation}
S.~Das~Choudhury, S.~Bashyam, Y.~Qiu, A.~Samal, T.~Awada, Holistic and
  component plant phenotyping using temporal image sequence, Plant Methods
  14~(1) (2018) 35.

\bibitem{nakano_inferD0Lerrorfree}
R.~Nakano, N.~Yamada, Number theory-based induction of deterministic
  context-free {L}-system grammar, in: International Conference on Knowledge
  Discovery and Information Retrieval, SCITEPRESS, 2010, pp. 194--199.

\bibitem{runqiang_inferGA}
B.~Runqiang, P.~Chen, K.~Burrage, J.~Hanan, P.~Room, J.~Belward, Derivation of
  {L}-system models from measurements of biological branching structures using
  genetic algorithms, in: Proceedings of the International Conference on
  Industrial, Engineering and Other Applications of Applied Intelligent
  Systems, Springer, 2002, pp. 514--524.

\bibitem{bernard_pmitdv3}
J.~Bernard, I.~McQuillan, New techniques for inferring {L}-systems using
  genetic algorithm, in: Proceedings of the 8th International Conference on
  Bioinspired Optimization Methods and Applications, Vol. 10835 of Lecture
  Notes in Computer Science, Springer, 2018, pp. 13--25.

\bibitem{bernard_pmitml}
J.~Bernard, I.~McQuillan, A fast and reliable hybrid approach for inferring
  {L}-systems, in: Proceedings of the 2018 International Conference on
  Artificial Life, MIT Press, 2018, pp. 444--451.

\bibitem{prusinkiewicz1990visualization}
P.~Prusinkiewicz, J.~Hanan, Visualization of botanical structures and processes
  using parametric {L}-systems, in: Scientific visualization and graphics
  simulation, John Wiley \& Sons, Inc., 1990, pp. 183--201.

\bibitem{back_geneticalgorithm}
T.~B\"ack, Evolutionary Algorithms in Theory and Practice: Evolution
  Strategies, Evolutionary Orogramming, Genetic Algorithms, Oxford {U}niversity
  {P}ress, 1996.

\bibitem{ben_naoum_surveryLsystems}
F.~Ben-Naoum, A survey on {L}-system inference, INFOCOMP Journal of Computer
  Science 8~(3) (2009) 29--39.

\bibitem{doucet_algebra}
P.~Doucet, The syntactic inference problem for {D0L}-sequences, L Systems
  (1974) 146--161.

\bibitem{mcquillan_poly}
I.~McQuillan, J.~Bernard, P.~Prusinkiewicz, Algorithms for inferring
  context-sensitive {L}-systems, in: 17th International Conference on
  Unconventional Computation and Natural Computation, Vol. 10867 of Lecture
  Notes in Computer Science, Springer, 2018, pp. 117--130.

\bibitem{hyperparameter}
J.~Bergstra, Y.~Bengio, Random search for hyper-parameter optimization, Journal
  of Machine Learning Research 13 (2012) 281--305.

\bibitem{grefenstette_optimalGAparameter}
J.~J. Grefenstette, Optimization of control parameters for genetic algorithms,
  IEEE Transactions on Systems, Man and Cybernetics 16~(1) (1986) 122--128.

\end{thebibliography}


\end{document}